%% file: main.tex
\documentclass[letterpaper, 10 pt, conference]{ieeeconf}  

\IEEEoverridecommandlockouts                              

\usepackage{graphicx}
\usepackage{caption}
\usepackage{subcaption}
\usepackage{amsmath, amssymb}
\usepackage{booktabs}
\usepackage{multirow} 
\usepackage{microtype}
\usepackage{xcolor}
\usepackage{siunitx}
\usepackage{wrapfig}
\usepackage{hyperref}
\usepackage{tabularx}
\usepackage{booktabs}
\usepackage{ragged2e}
\usepackage{pifont}
\usepackage{tikz}

\title{\LARGE \bf
VibeAct: Vibration to Actions for Contact-Rich Reactive Robot Dexterity
}

\author{Yuemin~Mao$^{*, 1}$,
Uksang~Yoo$^{*, 1}$,
Jean~Oh$^{1}$,
Jonathan Francis$^{1, 2}$,
Jeffrey~Ichnowski$^{1}$
\thanks{$^*$Equal contribution.}
\thanks{$^{1}$Robotics Institute, Carnegie Mellon University.}
\thanks{$^{2}$Bosch Center for Artificial Intelligence.}
}

\begin{document}
\raggedbottom

\newcommand{\method}{\textsc{VibeAct}}

\newcommand{\inlinecomment}[3]{\textcolor{#1}{[#2 -#3]}}
\newcommand{\yuemin}[1]{\inlinecomment{red}{#1}{YM}}

\IEEEaftertitletext{
    \begin{center}
    \includegraphics[width=0.9\textwidth]{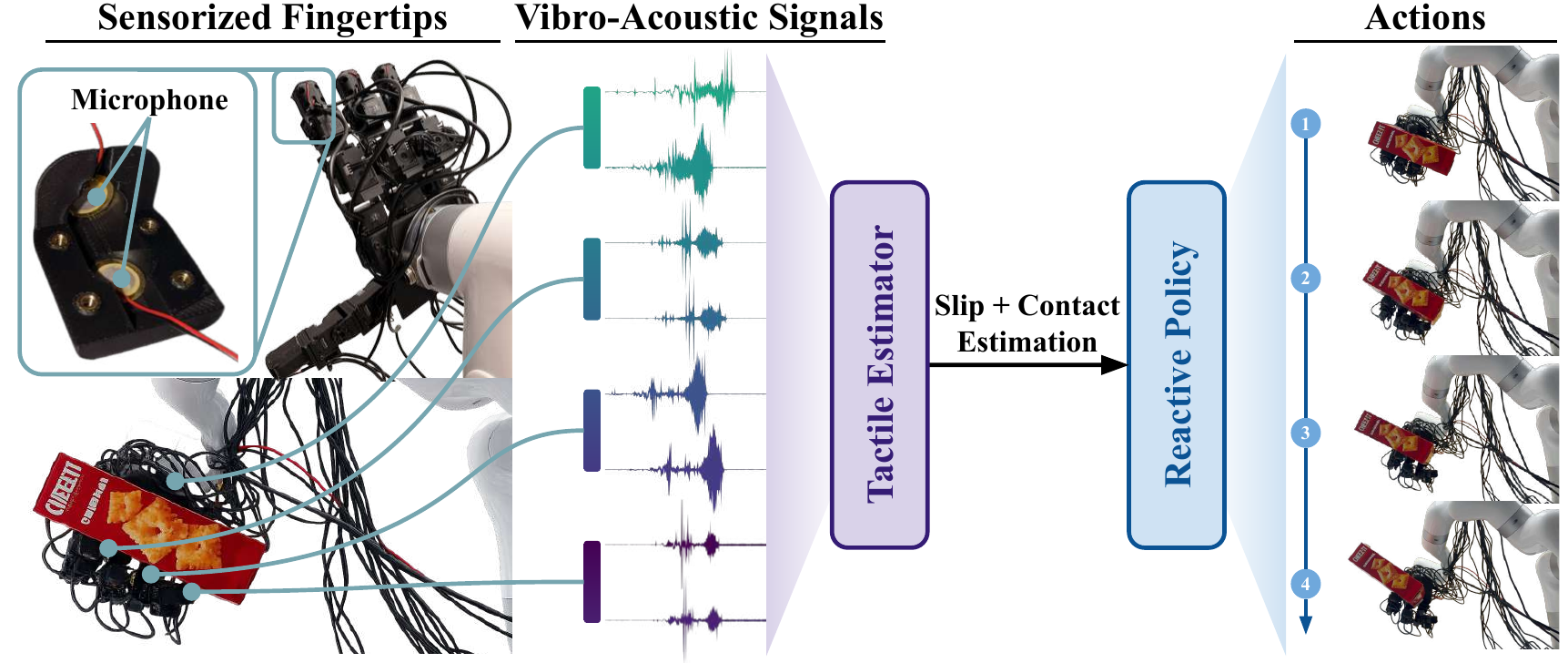}
    \captionof{figure}{\textbf{\method{}} connects real vibrotactile sensing to simulation-based policy learning through an explicit intermediate representation of contact and slip. A tactile estimator infers this representation from microphone signals, and the policy learns to act on the same representation in simulation.}
    \label{fig:front}
    \end{center}
}

\maketitle

\begin{abstract}
Dexterous manipulation depends on contact events that are fast, local, and often visually occluded. Piezoelectric microphones provide compact, high-bandwidth sensing of these interactions, but the resulting vibro-acoustic signals are difficult to simulate faithfully for end-to-end sim-to-real policy learning on dexterous robot hands. 
We propose \method{}, a framework that bridges real vibrotactile sensing and simulation-based reinforcement learning through a shared compact tactile representation. Guided by the microphones' sensitivity to contact transients and sliding-induced vibrations, we define this representation as binary contact onset, binary slip presence, and scalar slip magnitude for each fingertip. In the real world, we embed piezoelectric microphones in a dexterous robot hand and collect vibro-acoustic signals synchronized with robot and object states during teleoperation. We replay the recorded states in a calibrated digital clone to generate per-finger contact and slip labels for training a tactile estimator to infer the tactile representation from real microphone signals. In simulation, manipulation policies are trained using the same representation computed directly from simulated contact dynamics.
This decoupling lets policies exploit rapid tactile feedback without simulating raw audio. Across eight contact-rich manipulation tasks, \method{} improves average success from 21.6\% to 50.5\% over a baseline using proprioception and point clouds in simulation. The transferred policies also outperform the baseline on all three tasks evaluated on a physical dexterous hand-arm platform.
Project videos and additional details are at \href{https://vibeact.github.io/}{vibeact.github.io}.

\end{abstract}

\section{INTRODUCTION}
Open-world manipulation requires rapid responses to contact and slip events~\cite{nakamura2017complexities}. When a fingertip first makes contact, or when a grasp transitions from sticking to sliding, the relevant hand-object interaction cues are often brief, subtle, or visually occluded~\cite{dahiya2009tactile}. We present \method{}, a framework that uses piezoelectric microphones embedded in a robot hand to estimate these contact and slip interactions, incorporating the resulting estimates as tactile observations for learned dexterous manipulation policies (Fig.~\ref{fig:front}).


Piezoelectric microphones offer a promising tactile sensing modality for robot hands. Compared with popular vision-based tactile sensors such as GelSight~\cite{yuan2017gelsight} and Digit~\cite{lambeta2020digit}, they are inexpensive, compact, and high-bandwidth. Additionally, they can be mounted away from the contact surface while still capturing structure-borne vibrations generated by impacts, stick-slip motion, and sliding~\cite{mao2025hearing,yoo2024poe,lee2025sonicboom,yoo2026aslip}. The cost of these advantages is that the recorded waveforms depend not only on the underlying contact event but also on finger material, adhesive, mounting location, object texture, amplification, and motor-induced vibrations~\cite{clarke2022diffimpact}. These dependencies make vibro-acoustic signals difficult to simulate accurately enough for end-to-end policy learning.

Learning dexterous manipulation directly from real-world vibrotactile observations is also challenging. Collecting large-scale contact-rich demonstrations is labor-intensive, while online reinforcement learning is costly and potentially unsafe during early exploration. Together, these constraints motivate an approach that connects real vibrotactile sensing with simulation-based policy learning without audio simulation.



The central insight of this work is that a compact tactile representation grounded in the microphones' sensing capabilities can provide a shared interface between sensing and control for sim-to-real policy learning. Piezoelectric microphones are particularly well suited to capturing brief, high-amplitude vibrations at initial contact and friction-induced vibrations during sliding. To take advantage of these sensing strengths, we select binary contact onset, binary slip presence, and scalar slip magnitude for each fingertip as the \method{} tactile representation. These physically defined channels can be estimated from microphone recordings and computed directly from simulated contact dynamics. This correspondence allows us to evaluate both how accurately the representation can be recovered from real signals and how effectively it supports dexterous manipulation.

We instantiate \method{} on a dexterous hand-arm platform with piezoelectric microphones inside each fingertip. During teleoperated hand-object interactions, we record microphone signals synchronized with robot and object states. We replay the recorded trajectories in a calibrated digital clone to automatically generate per-finger contact and slip supervision for training the tactile estimator. Separately, we train PPO policies~\cite{schulman2017proximal} entirely in simulation using the same representation alongside proprioceptive and point-cloud observations. At deployment, the trained estimator replaces the simulator as the source of tactile observations.

We evaluate the effectiveness of the \method{} tactile representation for policy learning across eight tasks in simulation spanning regrasping, in-hand reorientation, and insertion. The best tactile configuration improves success rates over a baseline using proprioception and point clouds by up to 56.4 percentage points. Although the best combination of channels varies by task, the full representation improves performance on all eight tasks, achieving the highest average success rate at 50.5\%, compared with 21.6\% for the baseline. On three real-world tasks, the transferred policies using the full representation estimated from microphone signals achieve an average success rate of 65.0\%, compared with 26.7\% for the baseline.

In summary, this paper makes the following contributions:
\begin{itemize}
    \item A fingertip-integrated vibrotactile sensing system using piezoelectric microphones, paired with a digital-clone labeling pipeline that automatically generates per-finger contact and slip supervision from real-world recordings.
    \item A sim-to-real framework for dexterous manipulation using a shared compact tactile representation of binary contact onset, binary slip presence, and scalar slip magnitude, selected according to the sensing capabilities of the embedded microphones.
    \item An empirical study demonstrating the effectiveness of this representation across eight simulated manipulation tasks, identifying task-dependent benefits of its tactile channels, and validating its use through real-world deployment on three tasks.
\end{itemize}


\section{Related Work}

\method{} lies at the intersection of tactile sensing, vibrotactile perception, and tactile representation learning for robot manipulation. We review these areas with a focus on sensor integration, learning from vibration signals, and tactile observations for control.

\subsection{Tactile Sensing for Dexterous Manipulation}
Tactile sensing is critical for dexterous manipulation as many task-relevant events occur at the contact interface and are often difficult to infer from vision alone~\cite{niu2026learning, liu2025vitamin}. Prior work has explored a wide range of tactile sensing modalities for robotic manipulation. Vision-based tactile sensors have enabled high-resolution reconstruction of contact geometry~\cite{yuan2017gelsight, lambeta2020digit}, shear and slip~\cite{yuan2015measurement, dong2017improved}, surface texture~\cite{li2013textures, luo2018vitac}, and contact-rich manipulation behaviors~\cite{she2021cable, alspach2019soft, kim2022active, oller2023manipulation, lin2025lighttact}. However, scaling them to multi-fingered robot hands introduces challenges including mechanical bulk, camera bandwidth, illumination constraints, and substantial compute overhead. Magnetic tactile skins~\cite{bhirangi2021reskin,bhirangi2025anyskin,hellebrekers2020softmagnetic,tomo2018uskin} and capacitive or resistive sensors~\cite{huang2024vitac,liu2024capacitive,wistreich2025dexskin} can offer more compact designs and lower-dimensional observations, but often require changes to the fingertip structure or contact surface material.

Vibrotactile sensing complements these approaches with compact hardware that captures rapid contact dynamics with low computational overhead. In \method{}, fingertip-integrated piezoelectric microphones record contact-induced vibrations at 48 kHz without altering the external fingertip geometry or surface texture.

\subsection{Vibrotactile Sensing in Robotics}
Contact interactions generate rich structure-borne vibrations that encode information about contact states and interaction dynamics, motivating the use of vibrotactile sensing in robotics~\cite{lu2023active, rupavatharm2023sonicfinger}. Prior work has leveraged vibrotactile sensing for tasks including material property estimation~\cite{yi2026sound}, contact estimation~\cite{lee2025sonicboom, zhang2025vibecheck}, object and action recognition~\cite{gandhi2020swoosh, liu2025sonicsense}, granular flow estimation~\cite{clarke18granular}, tool use~\cite{zhang2019leveraging}, and slip estimation~\cite{yoo2026aslip}. Beyond perception, a related line of work feeds audio-visual and vibrotactile signals directly into learned manipulation policies, typically through behavior cloning on paired observations and expert actions~\cite{mejia2024hearing, thankaraj2023sounds, du2022play, liu2024maniwav, li2023see}. By training entirely in the real world, these methods avoid simulating vibro-acoustic signals, but they depend on expert demonstrations and have not been extended to dexterous multi-fingered hands.

Our goal is to scale policy learning with embedded vibrotactile sensing without collecting expert demonstrations. Reinforcement learning in physics simulation offers this scalability~\cite{qi2025simple, xing2025stabilizing}, but training directly on vibro-acoustic signals would require realistic microphone observations generated through accurate modeling of material properties, structural resonances, and actuator noise~\cite{clarke2022diffimpact}. \method{} avoids this requirement by using real recordings to train a tactile estimator and training manipulation policies entirely in simulation.

\subsection{Tactile Representations for Policy Learning}
Prior work has connected tactile sensing and simulation-based policy learning through representations with varying degrees of physical interpretability. Explicit representations include binary contact~\cite{yin2023rotating}, spatial contact information~\cite{qi2023general, yuan2024synesthesia}, and center of pressure~\cite{pan2026beyond}, with observations obtainable without training a perception model. Learned latent representations offer another route by distilling privileged policy features into tactile state estimators~\cite{chen2026ptld} or aligning real tactile embeddings with simulator-native observations~\cite{yang2026sblr}, without prescribing a physical meaning for each feature dimension.


\method{} uses an explicit tactile representation of binary contact onset, binary slip presence, and scalar slip magnitude for each fingertip. This choice reflects the embedded microphones' strengths in sensing contact transients and sliding-induced vibrations. Each channel can be estimated from microphone signals and computed from simulated contact dynamics, providing a shared observation space for real-world sensing and simulation-based policy learning.

\section{Problem Formulation}
\label{sec:problem_formulation}
We consider a dexterous hand manipulating rigid objects under partial observability. At each control step $t$, the policy observes proprioception $q_t$, a point cloud $p_t$, and a tactile representation $z_t$. It outputs a continuous action $a_t$ for the arm and hand joints. The goal is to maximize task return for contact-rich manipulation tasks. 

\input{floats/fig_setup}

The tactile hardware consists of piezoelectric microphones embedded in each robot fingertip (Fig. \ref{fig:setup}). At time $t$, the microphones produce an $n$-channel vibration signal $x_{t-w:t} \in \mathbb{R}^{n \times W}$ over a temporal window of length $W$. Directly training policies in simulation on $x$ would require accurate simulation of vibration propagation and sensor transfer functions. We instead define a compact tactile representation:%
\begin{equation}
\label{eq:contact_repr}
z_t = \{z_t^i\}_{i=1}^{4}, \qquad
z_t^i = [b_t^i, m_t^i, e_t^i],
\end{equation}%
where $b_t^i$ denotes a binary slip state, $m_t^i$ denotes scalar slip magnitude, and $e_t^i$ denotes a sparse contact-onset event for finger $i$. The real-world perception problem is to learn a mapping $f_\theta : x_{t-w:t} \mapsto z_t$ from vibro-acoustic signals to tactile representation. The control problem is to learn a manipulation policy $\pi_\phi(a_t \mid q_t, p_t, z_t)$ in simulation, where $z_t$ is computed directly from contact dynamics.

\section{Method}
\method{} has four components. First, vibrotactile sensing hardware with piezoelectric microphones embedded in fingertips captures contact-induced vibrations. Second, a data collection and digital-clone labeling pipeline pairs these recordings with contact and slip labels. Third, a tactile estimator learns to infer the compact tactile representation from microphone signals using these labels. Fourth, manipulation policies are trained using the same representation computed from simulated contact dynamics. At deployment, the estimator supplies the policies with tactile observations from real microphone signals.

\input{floats/fig_architecture}

\subsection{Vibrotactile Hardware}
The hardware setup consists of an xArm7 and a LEAP hand~\cite{shaw2023leaphand}. We embed two piezoelectric microphones along orthogonal internal surfaces of each fingertip, preserving its original external geometry and contact surface (Fig.~\ref{fig:front}, left). Vibrations transmitted through the fingertip deform the microphones' piezoelectric elements, generating electrical signals that capture contact-onset transients and friction-induced vibrations during sliding. The orthogonal mounting orientations provide complementary measurements of these vibrations. An audio mixer synchronizes all eight channels across the four fingertips (Fig.~\ref{fig:setup}).

\subsection{Real-World Data Collection and Digital-Clone Labeling}
During data collection, we teleoperate the robot to interact with objects while recording microphone signals at 48\,kHz, arm and hand joint positions and velocities at 30\,Hz, and synchronized timestamps. Objects are either rigidly fixed to the table or freely moving and tracked by a mocap system. The embedded piezoelectric microphones primarily capture structure-borne vibrations, so we leave ambient acoustic noise uncontrolled during data collection but account for vibrations generated by robot actuation. Specifically, we include free-space arm and finger motions as negative examples to help the tactile estimator distinguish actuation-induced vibrations from contact events.

We replay the recorded robot and object states in a calibrated digital clone in MuJoCo~\cite{todorov2012mujoco} to obtain contact information aligned with the microphone recordings (Fig.~\ref{fig:data_labeling}). At each timestep, the simulator provides the location, normal force, and relative tangential velocity of each active fingertip contact. We use this information to generate supervision for the tactile representation defined in Sec.~\ref{sec:contact-slip} without manual annotation. We validate the resulting labels in Sec.~\ref{sec:digital-clone}.




\input{floats/fig_data_labeling}

\subsection{Tactile Representation}
\label{sec:contact-slip}
The digital clone provides a broad range of contact information, from which we select features particularly well suited to the piezoelectric microphones' sensing capabilities. We retain binary contact onset, binary slip presence, and scalar slip magnitude for each fingertip, reflecting the microphones' sensitivity to contact transients and sliding-induced vibrations. We omit contact location, surface normal, tangential direction, and force vectors to keep the representation compact.

For fingertip $i$ in Eq.~\ref{eq:contact_repr}, contact onset $e_t^i$ is a one-step binary pulse marking a new contact. When multiple contacts occur on the same fingertip, we mark onset whenever any new contact appears. To compute the slip channels, we take the maximum tangential slip speed across active contacts, yielding a single speed per fingertip. We filter the simulator-derived speed to suppress artifacts before computing the slip labels. Values above $50\,\mathrm{mm/s}$ are replaced with the median of the five previous valid samples, followed by a five-frame causal median filter and an exponential moving average with $\alpha=0.15$. Slip presence $b_t^i$ indicates whether the filtered speed $\lVert v_t^i\rVert$ exceeds the threshold $\lVert v_{\mathrm{slip}}\rVert=5\,\mathrm{mm/s}$, and slip magnitude $m_t^i=\lVert v_t^i\rVert$ when slipping and zero otherwise. Together, these channels capture the transient nature of contact onset and the persistence of sliding.


\subsection{Tactile Estimator}

We train a tactile estimator that maps real microphone signals to tactile representations generated by the digital-clone data labeling pipeline, assuming that contact and slip retain the same physical meaning in both simulation and real-world vibro-acoustic sensing.

\paragraph{Network}
The estimator consists of four independent per-finger subnetworks (Fig.~\ref{fig:architecture}, left). For finger $i$, the input tensor $\mathbf{X}^{i} \in \mathbb{R}^{n \times M \times T}$ contains multi-channel log-mel spectrograms computed over 200\,ms windows from $n=2$ synchronized microphones, with $M$ mel bins and $T$ temporal frames. We use absolute dB magnitudes and dataset-wide normalization to preserve amplitude information while compensating for microphone gain differences. A learnable microphone-gating layer first suppresses noisy channels, after which the gated spectrograms are processed by a convolutional encoder with frequency-only pooling to preserve temporal resolution for transient events. Temporal convolutions and attention pooling produce per-microphone embeddings, fused into a per-finger representation. Separate heads then predict contact-onset probability, slip-presence probability, and slip magnitude, with the magnitude head additionally conditioned on amplitude statistics.

\paragraph{Training objectives}
We train the estimator with a multi-task objective over contact onset, slip presence, and slip magnitude. Let $y^{i}_{\text{onset}}, y^{i}_{\text{slip}} \in \{0,1\}$ denote the binary contact-onset and slip labels, and $\|v^{*\,i}\|$ the ground-truth tangential slip magnitude from digital-clone replay. We supervise the binary tasks with class-weighted binary cross-entropy losses
$\mathcal{L}^{i}_{\text{onset}} =
\mathrm{BCE}_{w_{+}^{\text{onset}}}
\!\left(p^{i}_{\text{onset}}, y^{i}_{\text{onset}}\right)$ and $
\mathcal{L}^{i}_{\text{slip}} =
\mathrm{BCE}_{w_{+}^{\text{slip}}}
\!\left(p^{i}_{\text{slip}}, y^{i}_{\text{slip}}\right)$, 
where $w_{+}^{\text{onset}}>1$ compensates for sparse contact-onset events and $w_{+}^{\text{slip}}<1$ balances the predominance of slip-positive frames. We supervise slip magnitude with a Huber loss only on slip-positive frames:
\[
\mathcal{L}^{i}_{\text{mag}} = \mathbf{1}[\|v^{*\,i}\| > \|v_{\mathrm{slip}}\|] \cdot \mathrm{Huber}_{\delta} \!\left(\|\hat{v}^{i}\|,\|v^{*\,i}\|\right),
\]
where $\|v_{\mathrm{slip}}\|=5\,\mathrm{mm/s}$ is the slip threshold used for labeling and $\|\hat{v}^{i}\|$ is the predicted slip magnitude. The overall training objective averages losses across all four fingers: %
\[
\mathcal{L} =\frac{1}{4}\sum_{i=1}^{4}\left(\lambda_{\text{onset}}\mathcal{L}^{i}_{\text{onset}}+\lambda_{\text{slip}}\mathcal{L}^{i}_{\text{slip}}+\lambda_{\text{mag}}\mathcal{L}^{i}_{\text{mag}}\right).
\]%

\subsection{Policy Learning}
\label{sec:policy-learning}

We use the same policy architecture across tasks in MuJoCo~\cite{todorov2012mujoco}, trained with PPO~\cite{schulman2017proximal}, varying only the scene and task reward. All tasks use the same xArm7 and LEAP hand model, under per-episode domain randomization of object friction, mass, and pose, so the policy must rely on contact feedback rather than memorized geometry.

At each step, the policy receives a tuple $o_t = (p_t,\, q_t,\, z_t)$, where $p_t$ is a fixed-camera point cloud, $q_t$ is proprioception over the hand joints and, where applicable, the arm pose, and $z_t \in \mathbb{R}^{12}$ is the contact-and-slip vector from Sec.~\ref{sec:contact-slip} (Fig.~\ref{fig:architecture}, right). A PointNet-style branch~\cite{qi2017pointnet} maps $p_t$ to a permutation-invariant geometric feature, while separate MLP branches encode $q_t$ and $z_t$, and the three features are concatenated and passed to symmetric actor and critic heads. Each task combines a dense, task-specific progress term $r^{\text{task}}_t$ with a small set of auxiliary terms weighted identically across all tasks, giving $r_t = r^{\text{task}}_t - \lambda_{\text{smooth}} \lVert a_t - a_{t-1} \rVert^2 - \lambda_{\text{drop}}\, d_t + \lambda_{\text{succ}}\, s_t$, where $d_t, s_t \in \{0,1\}$ indicate a drop and a successful completion at step $t$, so the auxiliary terms penalize abrupt action changes, discourage drops, and reward task completion. The progress term $r^{\text{task}}_t$ is the negative distance to a goal pose for the grasp, lift, place, and insert tasks, and a signed rotation and upward-displacement rate for the in-hand reposition and climb tasks. We hold $\lambda_{\text{smooth}}$, $\lambda_{\text{drop}}$, and $\lambda_{\text{succ}}$ fixed across all ablation conditions, so that only the policy's tactile observation $z_t$ varies between them.

\section{Experiments}

We evaluate \method{} from label generation through real-world manipulation. We first validate digital-clone supervision (Sec.~\ref{sec:digital-clone}), then assess the tactile estimator's accuracy on real microphone signals (Sec.~\ref{sec:estimator-eval}). Finally, we examine whether the resulting tactile representation improves manipulation policies in simulation and whether the benefits transfer to the real world (Sec.~\ref{sec:task-performance}). Together, these evaluations test whether the \method{} tactile representation can bridge real vibrotactile sensing and simulation-based policy learning.

\subsection{Digital-Clone Label Validation} 
\label{sec:digital-clone}
\input{floats/tab_label_validation}
Our data collection considers two object settings: fixed objects, which are rigidly mounted to the table, and moving objects, which are tracked by mocap during interaction. To align the physical setup with the digital clone, we calibrate each fixed object's pose relative to the robot base using 50 distinct hand-object contact poses, and calibrate the mocap frame to the robot base using 120 poses for moving objects, achieving held-out alignment errors of 2.11\,mm and 1.01$^\circ$.

Contact-onset labels are derived from simulation and cross-checked against acoustic transients during labeling, so we focus primarily on validating slip presence and magnitude against measured physical ground truth in both settings. For fixed objects, \textit{Surface Sweep} evaluates slip against commanded Cartesian end-effector translation as we teleoperate the robot to sweep the fingertips along the object's surface while maintaining contact and a fixed hand configuration. For mocap-tracked objects, \textit{Stable Grasp} tests for spurious contact-onset and slip predictions as the arm moves while securely holding an object, and \textit{Probe} evaluates slip against mocap-measured object translation as an externally perturbed object moves within a stationary loose grasp. 

Table~\ref{tab:label_validation} shows contact-onset false-positive rates no greater than 0.25\% across all three conditions and slip-presence F1 scores of 0.997 and 0.964 for \textit{Surface Sweep} and \textit{Probe}, respectively. Although \textit{Stable Grasp} exhibits a higher slip false-positive rate, slip-magnitude MAE remains below 2\,mm/s across all three conditions, indicating small average errors in continuous speed estimates despite discrepancies in binary slip classification.

\subsection{Tactile Estimator Evaluation}
\label{sec:estimator-eval}
\input{floats/tab_acoustic_ablation}
We collect one teleoperated dataset under each object setting to train and evaluate the tactile estimator before policy deployment. The fixed-object dataset (five hours) efficiently scales diverse contact and sliding events, with slip induced purely by robot motion, but differs from the dynamics of dexterous manipulation. The moving-object dataset (under two hours) captures more realistic in-hand interactions, where slip can arise from finger motion, arm motion, and object dynamics. We pretrain on the fixed-object data to learn general vibro-acoustic patterns, then fine-tune on the moving-object data to match deployment. We evaluate all models on a held-out moving-object split, reporting F1 for binary contact-onset and slip-presence detection and mean absolute error (MAE) for slip magnitude, averaged over the four fingers.

\method{} tactile estimator achieves a contact-onset F1 of 0.597, slip-presence F1 of 0.913, and slip-magnitude MAE of 4.736\,mm/s (Table~\ref{tab:acoustic_ablation}). To account for class imbalance, we additionally report Area Under the Precision-Recall Curve (AUPRC), obtaining 0.627 for contact onset and 0.957 for slip presence, compared with positive-class prevalence baselines of 0.063 and 0.592, respectively. Because contact onset is substantially sparser and more temporally precise than persistent contact, we further train a contact-presence predictor, which achieves 0.965 F1 and 0.992 AUPRC compared with a positive-class prevalence baseline of 0.858. These results show that \method{} reliably detects contact and slip states, while precisely identifying sparse contact-onset events remains more challenging.

Table~\ref{tab:acoustic_ablation} compares training strategies and architectural variants. Pretraining alone underperforms \method{}, lowering contact-onset F1 by 35.7\% and slip-presence F1 by 14.5\% while raising slip-magnitude MAE by 35.5\%. This suggests a large domain gap between fixed-object and in-hand slip. Joint training on both datasets narrows this gap and marginally exceeds \method{} on slip-presence F1, but still underperforms overall, reducing contact-onset F1 by 7.5\% and increasing slip-magnitude MAE by 3.8\%. We attribute this to the differing temporal structure: contact onset is a sparse transient requiring precise temporal alignment, while slip presence is temporally persistent and transfers more readily across domains. \method{}'s pretrain-then-fine-tune strategy thus preserves transient sensitivity while retaining strong slip estimation.

We further evaluate parameter sharing across fingers. Sharing the encoder while keeping independent prediction heads reduces contact-onset F1 by 2.3\% and raises slip-magnitude MAE by 1.9\% relative to \method{}. Sharing both the encoder and prediction heads further compounds the degradation, reducing contact-onset F1 by 3.7\% and increasing MAE by 4.8\%. These results suggest that each fingertip exhibits distinct contact geometry and vibration propagation characteristics that benefit from dedicated feature extraction and prediction heads.

\subsection{Tasks}
\input{floats/tab_policy_performance}
\input{floats/fig_policy_performance}

We evaluate \method{} on eight contact-rich tasks spanning regrasping, in-hand reorientation, and insertion (Fig.~\ref{fig:tasks}). \textbf{Box Feed} and \textbf{Can Retract} fix the arm and use finger gaiting to translate a held YCB object~\cite{calli2015ycb} along its longitudinal axis in opposite directions. Wrist orientation is randomized while each task's motion direction remains fixed relative to the hand. \textbf{Cube Reorientation} and \textbf{Phone Flip} require in-hand reorientation along orthogonal axes. \textbf{Nut Rotation} and \textbf{Screwdriver Turning} require sustained rotational manipulation with coordinated hand-arm motion. \textbf{Peg Insertion} starts from a pregrasped cylinder and requires sideways insertion through post-contact alignment. \textbf{Card Pickup} requires grasping and lifting a thin card from a supporting surface.

\subsection{Task Performance}
\label{sec:task-performance}
We evaluate \method{} against a proprioception and point-cloud baseline (Prop+PC) with no tactile input, and ablate the tactile representation by incrementally adding contact onset, slip presence, and slip magnitude channels. Table~\ref{tab:policy-results} reports success rates averaged over three random seeds, with 100 evaluation trials per seed, while the policy training curves are shown in Fig.~\ref{fig:plots}.



\input{floats/fig_tab_tasks}


The best tactile configuration improves Card Pickup and Peg Insertion over the baseline by approximately $+56$ and $+51$ percentage points, respectively. It also improves Screwdriver Turning by $+39$ points, Phone Flip by $+38$ points, Cube Reorientation by $+31$ points, Box Feed by $+26$ points, Nut Rotation by $+19$ points, and Can Retract by $+16$ points. These gains span all eight tasks, supporting the value of tactile observations across diverse manipulation behaviors.

The full \method{} representation achieves the highest average success rate of 50.5\%, compared with 21.6\% for the baseline, and performs best on four of the eight tasks. Adding slip magnitude produces particularly large gains on Can Retract and Nut Rotation, highlighting the value of graded slip feedback for these tasks. However, the full representation is not optimal for every task: contact onset alone performs best on Cube Reorientation and Screwdriver Turning, while the combination of contact onset and slip presence performs best on Phone Flip and Card Pickup. These results suggest that the tactile channels play complementary, task-dependent roles, rather than following a uniform ranking of importance across tasks. Importantly, all tactile ablations in Table~\ref{tab:policy-results} are variants of \method{}, differing only in which components of the shared contact-and-slip representation are provided to the policy. Nevertheless, the full \method{} representation outperforms the no-tactile policy on all eight tasks, demonstrating the overall benefit of incorporating vibrotactile feedback.

We evaluate real-world transfer on three tasks spanning in-hand dexterity and hand-arm coordination. We deploy the simulation-trained policies directly on the physical platform, replacing the simulator-derived tactile channel with the trained estimator (Table~\ref{tab:real-world}). \method{} improves over the baseline with only proprioceptive and point-cloud observations on every deployed task. These results indicate that the contact-and-slip representation, estimated entirely from real microphone recordings and used as a drop-in observation channel, retains its benefit under real sensing and actuation noise.



\section{Conclusion}
We present \method{}, a framework for leveraging vibrotactile sensing in contact-rich dexterous manipulation without requiring audio simulation. By defining a compact contact-and-slip tactile representation that can be estimated from real piezoelectric microphone signals and computed from simulated contact dynamics, \method{} decouples the sensor problem from the control problem. A tactile estimator trained on real microphone signals with digital-clone-derived labels bridges the two domains, enabling policies trained entirely in simulation to exploit tactile feedback at deployment. Across eight contact-rich manipulation tasks, \method{} consistently outperforms a proprioception and point cloud baseline, with the largest gains on tasks requiring sustained reactive control such as in-hand rotation and peg insertion. These results suggest that physically grounded intermediate representations are a practical and effective route for incorporating high-bandwidth tactile sensing into sim-to-real policy learning.

A limitation and direction for future work is that \method{} uses a fixed tactile representation across tasks. The ablation results suggest that no single representation and variant of \method{} is uniformly optimal. Graded slip feedback is particularly useful for sustained contact-rich control, whereas discrete contact-onset and slip cues can be more informative for behaviors governed by distinct contact transitions. This suggests that the appropriate tactile abstraction is inherently task-dependent. Future work could explore task-conditioned or adaptive representations that selectively incorporate contact events, slip, force, contact location, or richer vibrotactile features according to the manipulation objective, while preserving a representation that can be reliably aligned between simulation and the real world.

\section*{ACKNOWLEDGMENT}
This work was supported by Samsung Research America and NSF Graduate Research Fellowship under Grant No. DGE2140739.
\bibliographystyle{IEEEtran}
\bibliography{abbreviations, references}
\end{document}

%% file: floats/fig_setup.tex
\begin{figure}
    \centering
    \includegraphics[width=0.8\linewidth]{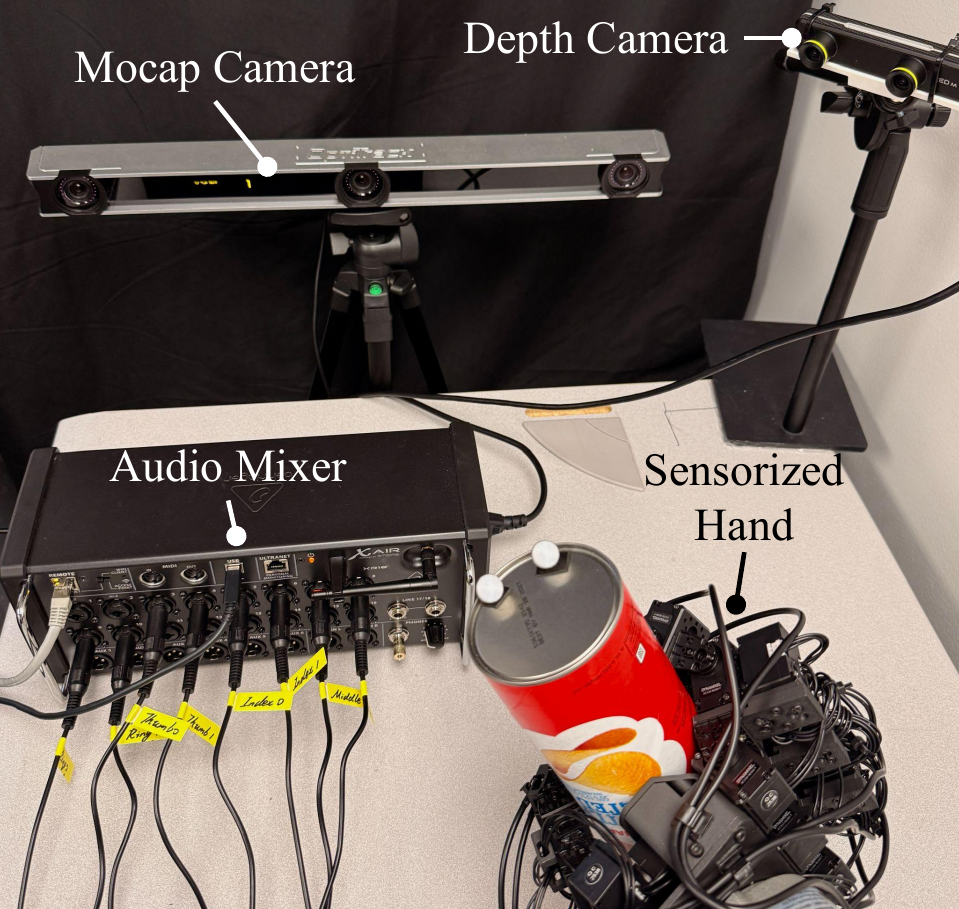}
    \caption{\textbf{Hardware setup.} The audio mixer synchronizes microphone signals; the mocap camera tracks object trajectories for digital-clone vibrotactile data labeling; the depth camera supplies point clouds during policy deployment.}
    \label{fig:setup}
    \vspace{-0.2cm}
\end{figure}

%% file: floats/fig_architecture.tex
\begin{figure*}[t]
  \centering
   \includegraphics[width=\textwidth]{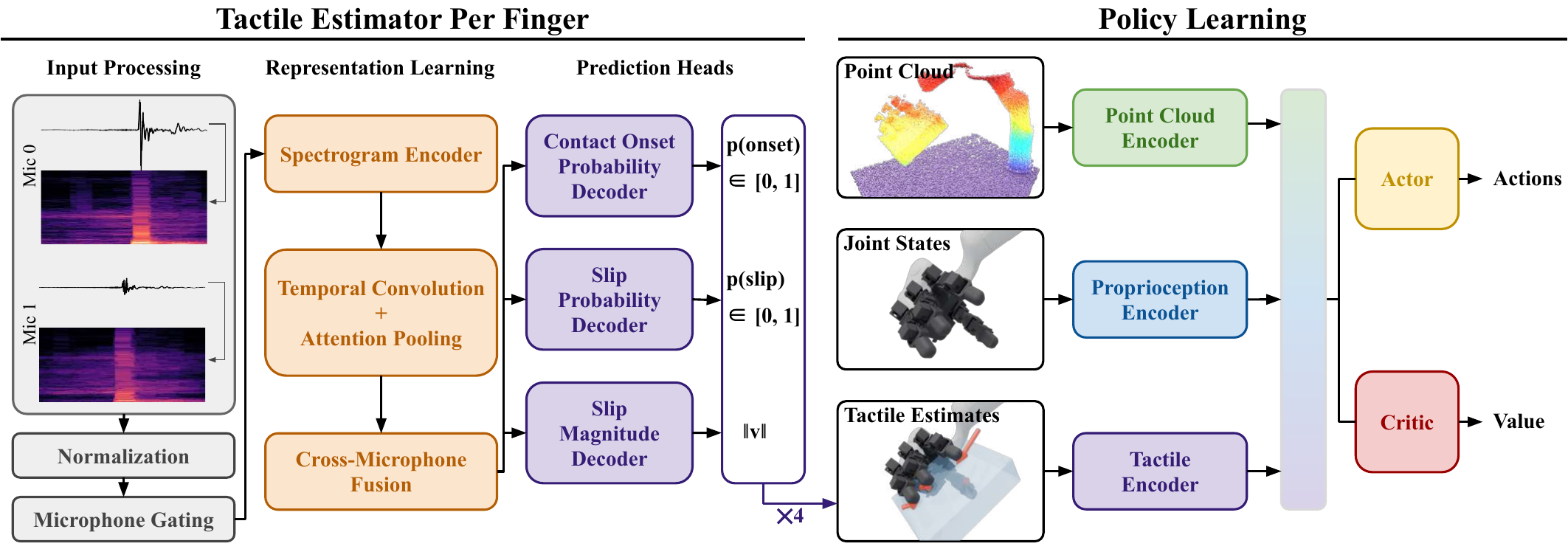}
  \caption{\textbf{An overview of \method{}.} We train a tactile estimator with four independent per-finger subnetworks using real-world data to map vibro-acoustic signals to a contact and slip representation. We then train RL policies that use this representation as an additional observation modality alongside point clouds and proprioception.} 
  \label{fig:architecture}
  \vspace{-0.2cm}
\end{figure*}

%% file: floats/fig_data_labeling.tex
\begin{figure}[t]
  \centering
  \includegraphics[width=\linewidth]{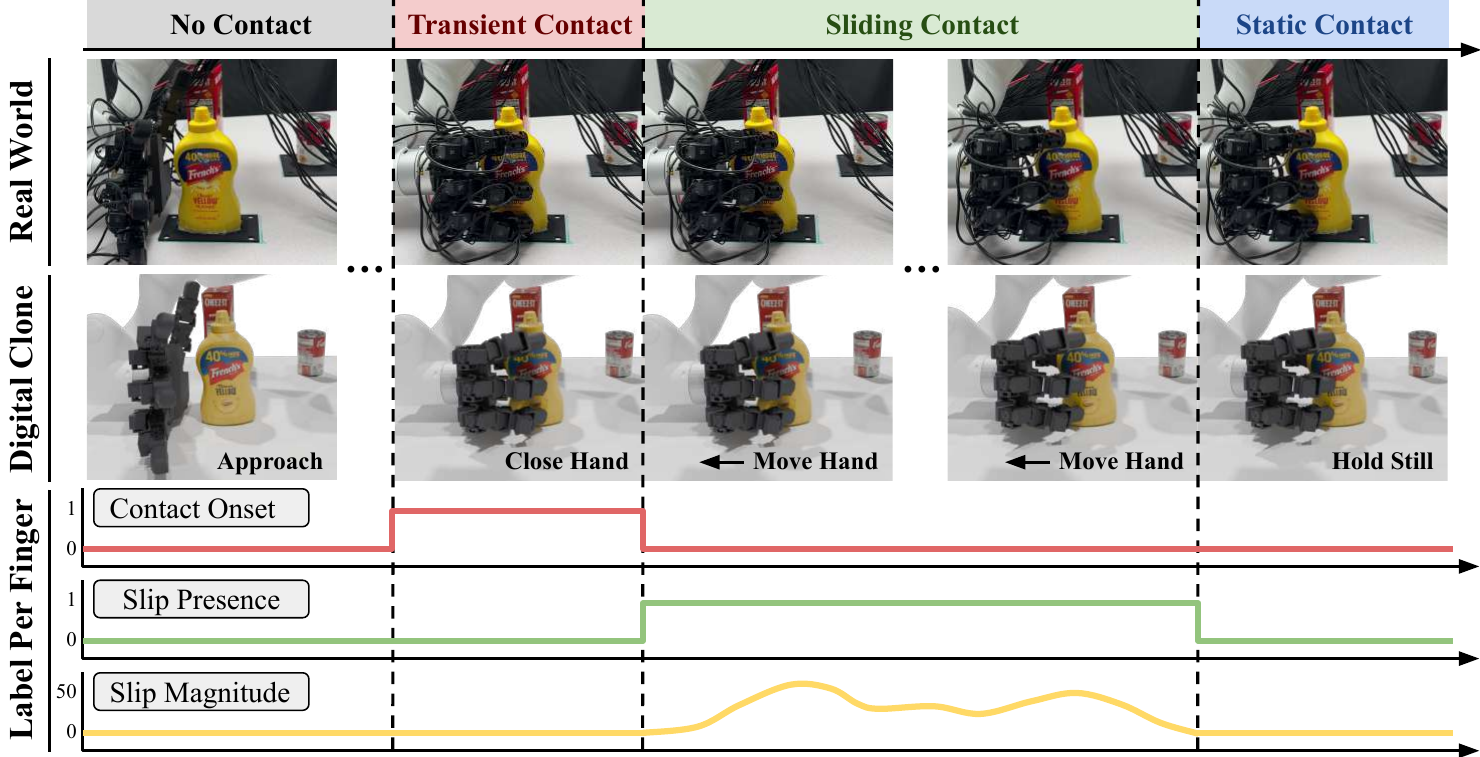}
  \caption{\textbf{Vibrotactile data labeling setup.} We replay real-world teleoperation recordings in a calibrated digital clone, whose contact solver generates per-finger contact and slip labels to train the tactile estimator without manual annotation.}
  \label{fig:data_labeling}
  \vspace{-0.2cm}
\end{figure}

%% file: floats/tab_label_validation.tex
\begin{table}[t]
  \centering
  \caption{\textbf{Digital-clone label validation against physical ground truth.} FP/FN: false-positive/negative rates; MAE: mean absolute error.}
  \resizebox{\linewidth}{!}{%
  \scriptsize
  \begin{tabular}{@{}lrccccc@{}}
    \toprule
    \textbf{Setting} &
    \textbf{Frames} &
    \textbf{Contact Onset} &
    \multicolumn{3}{c}{\textbf{Slip Presence}} &
    \textbf{Slip Magnitude} \\
    \cmidrule(lr){4-6}
    &
    $(\times 10^3)$ &
    \textbf{FP (\%)} &
    \textbf{FP (\%)} &
    \textbf{FN (\%)} &
    \textbf{F1} &
    \textbf{MAE (mm/s)} \\
    \midrule
    Surface Sweep &
    45 &
    0.02 &
    2.65 &
    0.43 &
    0.997 &
    $0.17 \pm 0.07$ \\
    Stable Grasp &
    36 &
    0.25 &
    20.44 &
    --- &
    --- &
    $1.86 \pm 1.41$ \\
    Probe &
    36 &
    0.11 &
    3.41 &
    3.78 &
    0.964 &
    $0.68 \pm 0.27$ \\
    \bottomrule
  \end{tabular}
  }
  \label{tab:label_validation}
  \vspace{-0.2cm}
\end{table}

%% file: floats/tab_acoustic_ablation.tex
\begin{table}[t]
  \centering
  \caption{
  \textbf{Tactile estimator performance.} Full \method{} uses sequential training with independent
  per-finger encoders and heads. Training strategy variants retain
  this architecture, while shared-encoder architecture variants retain
  sequential training. Results are averaged over all four fingers.}
  \label{tab:acoustic_ablation}
  \resizebox{\linewidth}{!}{%
  \begin{tabular}{@{}lccccc@{}}
    \toprule
    \shortstack{\textbf{Training}\\\textbf{Strategy}} &
    \shortstack{\textbf{Shared}\\\textbf{Encoder}} &
    \shortstack{\textbf{Shared}\\\textbf{Heads}} &
    \shortstack{\textbf{Contact Onset}\\\textbf{F1 $\uparrow$}} &
    \shortstack{\textbf{Slip Presence}\\\textbf{F1 $\uparrow$}} &
    \shortstack{\textbf{Slip Magnitude}\\\textbf{MAE (mm/s) $\downarrow$}} \\
    \midrule
    Pretraining only & \ding{55} & \ding{55} &
    $0.384 \pm 0.110$ &
    $0.781 \pm 0.060$ &
    $6.417 \pm 1.399$ \\
    Joint & \ding{55} & \ding{55} &
    $0.552 \pm 0.114$ &
    $\mathbf{0.923 \pm 0.036}$ &
    $4.914 \pm 0.792$ \\
    \midrule[0.4pt]
    Sequential & \ding{51} & \ding{51} &
    $0.575 \pm 0.077$ &
    $0.891 \pm 0.053$ &
    $4.964 \pm 0.838$ \\
    Sequential & \ding{51} & \ding{55} &
    $0.583 \pm 0.090$ &
    $0.907 \pm 0.049$ &
    $4.827 \pm 0.728$ \\
    \midrule[0.4pt]
    \multicolumn{6}{@{}l}{\textbf{\textit{Full \method{}}}} \\
    Sequential & \ding{55} & \ding{55} &
    $\mathbf{0.597 \pm 0.101}$ &
    $0.913 \pm 0.054$ &
    $\mathbf{4.736 \pm 0.658}$ \\
    \bottomrule
  \end{tabular}%
  }
  \vspace{-0.2cm}
\end{table}

%% file: floats/tab_policy_performance.tex
\begin{table}[t]
\centering
\caption{\textbf{Average task success across tactile observation ablations and random seeds in simulation.}}
\label{tab:policy-results}
\resizebox{\linewidth}{!}{%
\begin{tabular}{l
  S[table-format=2.1]
  S[table-format=2.1]
  S[table-format=2.1]
  S[table-format=2.1]}
\toprule
\multirow{2}{*}{\textbf{Task}}
  & {\textbf{Prop + PC}}
  & {\textbf{+}\,\begin{tabular}[c]{@{}c@{}}
       \textbf{Contact}\\
       \textbf{Onset}
     \end{tabular}}
  & {\textbf{+}\,\begin{tabular}[c]{@{}c@{}}
       \textbf{Slip}\\
       \textbf{Presence}
     \end{tabular}}
  & {\textbf{+}\,\begin{tabular}[c]{@{}c@{}}
       \textbf{Slip}\\
       \textbf{Magnitude}
     \end{tabular}} \\
  & {\textbf{(No Tactile) (\%)}}
  & {\textbf{(\%)}}
  & {\textbf{(\%)}}
  & {\textbf{(Full \textsc{VibeAct}) (\%)}} \\
\midrule
Box Feed            & 18.3 & 20.7            & 42.7            & {\textbf{44.3}} \\
Can Retract         & 51.3 & 33.7          & 49.7            & {\textbf{67.3}} \\
Cube Reorientation  & 31.3  & {\textbf{62.7}}        &    53.3         &  52.7 \\
Phone Flip          & 31.7 & 62.7            & {\textbf{69.3}} & 51.7            \\
Nut Rotation        & 25.7 & 29.3            & 11.3          & {\textbf{45.0}} \\
Screwdriver Turning & 0.0  & {\textbf{39.3}} & 21.7            & 24.3            \\
Peg Insertion       & 12.3  & 9.7         & 55.3          & {\textbf{63.3}} \\
Card Pickup         & 2.3  & 9.0         & {\textbf{58.7}} & 55.3           \\
\bottomrule
\end{tabular}%
}
\vspace{-0.2cm}
\end{table}

%% file: floats/fig_policy_performance.tex
\begin{figure}[t]
  \centering
  \includegraphics[width=\linewidth]{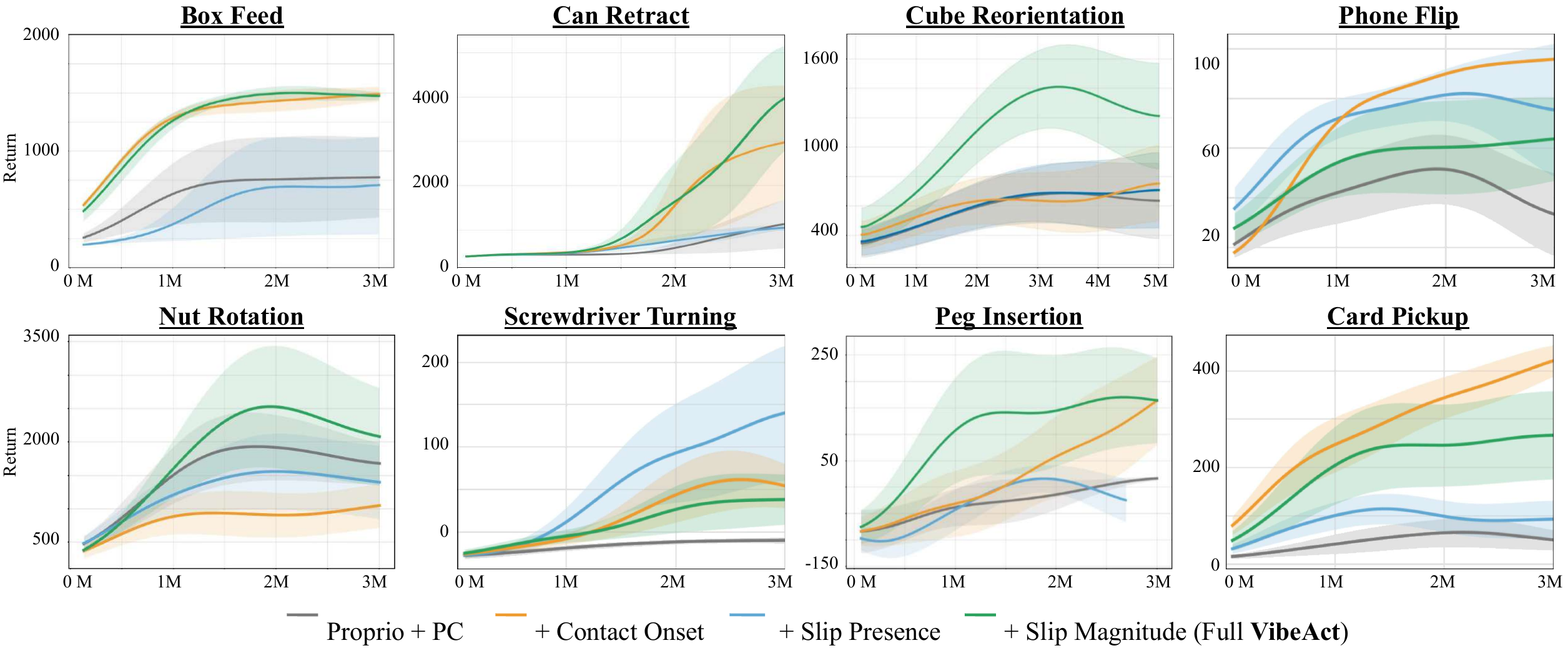}
  \caption{\textbf{Training curves for the \method{} policies and baselines.}}
  \label{fig:plots}
  \vspace{-0.2cm}
\end{figure}

%% file: floats/fig_tab_tasks.tex
\newsavebox{\deploymenttable}
\newlength{\deploymentwidth}

\setlength{\deploymentwidth}{0.27\textwidth}

\begin{figure*}[t]
\centering

\sbox{\deploymenttable}{%
  \scriptsize
  \setlength{\tabcolsep}{3pt}%
  \renewcommand{\arraystretch}{0.95}%
  \begin{tabular}{lcc}
    \toprule
    \textbf{Task} & \textbf{Prop+PC} & \textbf{\method} \\
    \midrule
    Box Feed
      & 4/20  & \textbf{12/20} \\
    Can Retract
      & 11/20 & \textbf{19/20} \\
    Nut Rotation
      & 1/20  & \textbf{8/20} \\
    \bottomrule
  \end{tabular}%
}

\begin{tikzpicture}

\node[inner sep=0] (tasks) {%
  \includegraphics[width=\textwidth]{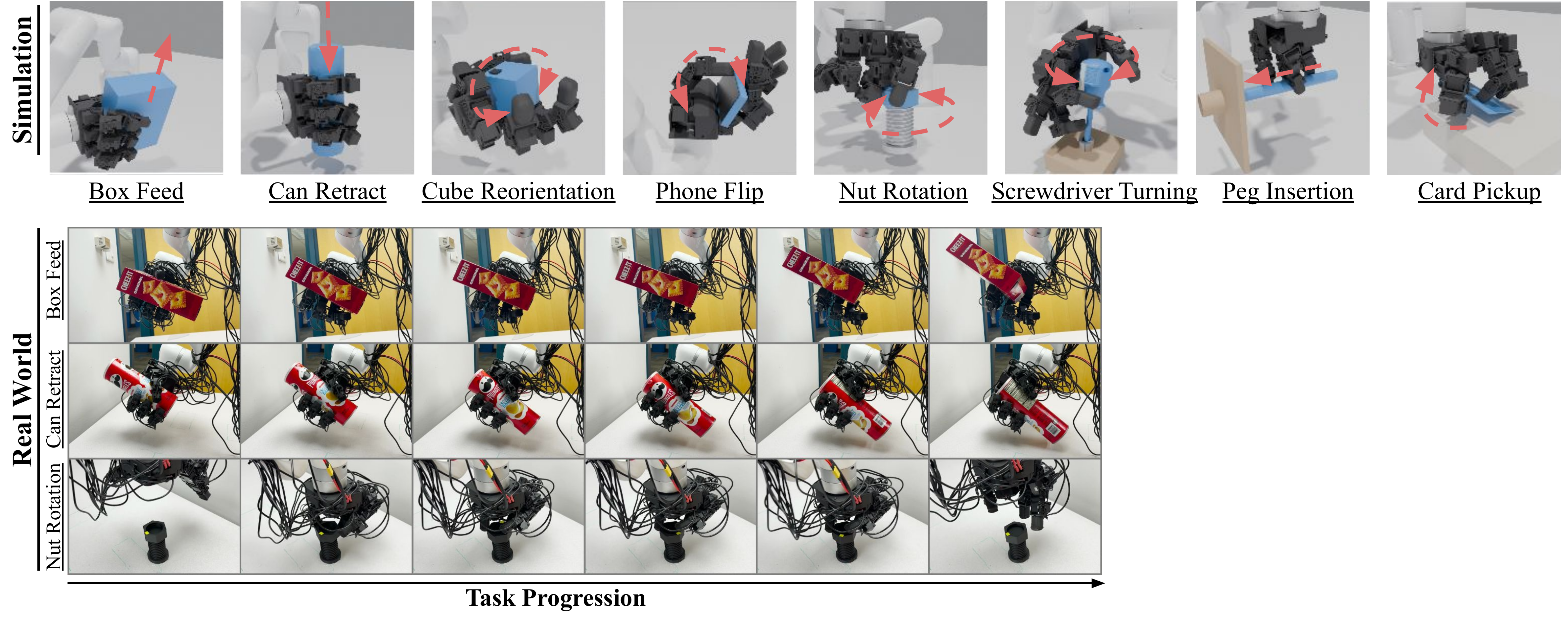}%
};

\node[
  anchor=south east,
  xshift=0mm,
  yshift=8mm,
  inner sep=5pt
] at (tasks.south east) {%
  \resizebox{\deploymentwidth}{!}{%
    \begin{minipage}{\wd\deploymenttable}

      \captionsetup{
        font=footnotesize,
        width=\linewidth,
        singlelinecheck=false,
        skip=2pt
      }

      \captionof{table}{%
        \textbf{Real-world deployment performance.}
        \textit{Prop+PC}: proprioception + point cloud baseline.
      }
      \label{tab:real-world}

      \centering
      \usebox{\deploymenttable}

    \end{minipage}%
  }%
};

\end{tikzpicture}

\caption{%
  \textbf{Tasks.}
  We evaluate \method{} on eight contact-rich tasks in simulation spanning regrasping, in-hand reorientation, arm-hand coordination, and insertion.
We deploy three of these tasks in the real world.
}
\label{fig:tasks}
\vspace{-0.2cm}

\end{figure*}